\documentclass[conference]{IEEEtran}
\usepackage{cite}
\usepackage{amsmath,amssymb,amsfonts}
\usepackage{algorithmic}
\usepackage{graphicx}
\usepackage{xcolor}
\usepackage{multirow}
\usepackage{booktabs}
\usepackage{ulem}
\def\BibTeX{{\rm B\kern-.05em{\sc i\kern-.025em b}\kern-.08em
    T\kern-.1667em\lower.7ex\hbox{E}\kern-.125emX}}
\begin{document}

\title{RecMind: LLM-Enhanced Graph Neural Networks for Personalized Consumer Recommendations}

\author{\IEEEauthorblockN{Chang Xue}
    \IEEEauthorblockA{
        \textit{Yeshiva University}\\
        cxue@mail.yu.edu}
    \and
    \IEEEauthorblockN{Youwei Lu}
    \IEEEauthorblockA{
        \textit{Oklahoma State University}\\
        youwei.lu@okstate.edu}
    \and
    \IEEEauthorblockN{Chen Yang}
    \IEEEauthorblockA{
        \textit{Case Western Reserve University}\\
        sophiayang1112@gmail.com}
    \and
    \IEEEauthorblockN{Jinming Xing}
    \IEEEauthorblockA{\textit{North Carolina State University} \\
        jxing6@ncsu.edu}
}

\maketitle

\begin{abstract}
    Personalization is a core capability across consumer technologies, streaming, shopping, wearables, and voice, yet it remains challenged by sparse interactions, fast content churn, and heterogeneous textual signals. We present RecMind, an LLM-enhanced graph recommender that treats the language model as a preference prior rather than a monolithic ranker. A frozen LLM equipped with lightweight adapters produces text-conditioned user/item embeddings from titles, attributes, and reviews; a LightGCN backbone learns collaborative embeddings from the user-item graph. We align the two views with a symmetric contrastive objective and fuse them via intra-layer gating, allowing language to dominate in cold/long-tail regimes and graph structure to stabilize rankings elsewhere. On Yelp and Amazon-Electronics, RecMind attains the best results on all eight reported metrics, with relative improvements up to +4.53\% (Recall@40) and +4.01\% (NDCG@40) over strong baselines. Ablations confirm both the necessity of cross-view alignment and the advantage of gating over late fusion and LLM-only variants.
\end{abstract}


\section{Introduction}
Personalized recommendation is now a first-class capability across consumer devices and services, from smart TVs and streaming boxes to shopping apps, wearables, and voice assistants. These settings magnify long-standing challenges in recommender systems: extremely sparse user-item interactions, rapid content churn, and heterogeneous signals (clicks, watches, purchases, reviews, and device-side contexts). Graph neural networks (GNNs) have become a strong backbone by modeling collaborative structure in user-item interaction graphs, yet they under-utilize the rich semantics available in product titles, descriptions, and user reviews. Large language models (LLMs), in contrast, excel at capturing such semantics and subtle preference cues but lack the inductive bias required to reason over collaborative patterns, and can be costly or unstable when used as monolithic recommenders. The key opportunity is to marry the structural strengths of GNNs with the semantic strengths of LLMs, without incurring heavy latency or risking language-only hallucinations.

We propose RecMind, an LLM-enhanced GNN for personalized consumer recommendations. RecMind treats the LLM as a preference prior rather than the end-to-end recommender: lightweight adapters condition a (frozen) LLM on item metadata and user textual signals to produce semantically grounded embeddings; a GNN simultaneously learns collaborative embeddings from the interaction graph. We align these two views through a joint contrastive objective that draws together language-derived and graph-derived representations of the same user/item while pushing apart mismatched pairs. At inference, RecMind fuses the aligned signals, via a simple, efficient gating mechanism within the GNN message-passing pathway, yielding rankings that are structurally consistent yet semantically aware. This design delivers three practical advantages for consumer technologies: (1) stronger generalization in cold-start and long-tail regimes, (2) robustness against language-only drift, and (3) deployability with modest compute by freezing the LLM and training only small adapters and the GNN.

Our contributions are as follows:
\begin{itemize}
    \item LLM-as-prior architecture for recsys. We introduce a principled integration where the LLM supplies semantic preference priors and the GNN preserves collaborative structure, tailored to consumer electronics use cases spanning screens, wearables, and voice.
    \item Contrastive alignment between language and interactions. A joint objective aligns text-conditioned and graph-conditioned embeddings, enabling consistent scoring across modalities and reducing over-reliance on either source.
    \item Parameter-efficient conditioning and fusion. Adapter-based LLM conditioning and a lightweight fusion gate make RecMind portable across backbones and friendly to real-time constraints.
    \item Comprehensive empirical study. Across diverse consumer datasets, RecMind improves top-K ranking metrics (e.g., NDCG@K/Recall@K), with ablations demonstrating gains in cold-start and robustness.
\end{itemize}

\section{Background and Related Work}
\textbf{GNN-based collaborative filtering.} Modern recommenders increasingly encode user-item graphs with GNNs \cite{wu2022graph, gao2022graph}. NGCF propagates embeddings over high-order connectivity to inject collaborative signals \cite{wang2019neural}; PinSage scales graph convolutions to web-scale, showing industrial effectiveness \cite{ying2018graph}; and LightGCN strips nonlinearities to a pure neighborhood aggregator that improves accuracy and efficiency \cite{he2020lightgcn}. Classical pairwise ranking with BPR remains a standard optimization for implicit feedback \cite{rendle2012bpr}. While these models capture structure well, they are text-agnostic and struggle in cold-start regimes where semantics matter \cite{kim2024large}.

\textbf{Sequential and contrastive recommenders.} Transformers advanced sequence modeling, improving next-item prediction from short histories \cite{vaswani2017attention, cheng2025unifying, zhang2025selective}. Self-supervised/contrastive objectives, e.g., S3-Rec \cite{zhou2020s3} (pre-training via mutual information) and CL4SRec \cite{xie2022contrastive}, further mitigate sparsity; later work (SimGCL/XSimGCL) showed that simple noise-based views rival graph augmentations \cite{liu2024simgcl,yu2023xsimgcl}. These techniques enrich representations but do not directly align textual semantics with collaborative structure.

\textbf{Multimodal-aware recommenders.} Review-driven models (DeepCoNN \cite{zheng2017joint}, NARRE \cite{chen2018neural}) demonstrate that textual signals improve rating prediction and explanations; news recommenders such as NAML \cite{wu2019neural} and multimodal GNNs (MMGCN \cite{wei2019mmgcn}) fuse titles/bodies/categories or video/image cues, respectively. However, these methods predate LLMs and typically rely on task-specific encoders rather than general semantic priors.

\textbf{LLMs for recommendation.} Recent studies probe LLMs as zero-shot rankers and instruction-following recommenders (e.g., GPT4Rec \cite{li2023gpt4rec}, InstructRec \cite{zhang2025recommendation}), and surveys chart generative/LLM-augmented paradigms \cite{hou2024large}. Evidence shows promise but also limitations: difficulty modeling interaction order and prompt-position/popularity biases; fairness concerns; and deployment costs when using LLMs as monolithic rankers \cite{zhao2024recommender}. Hybrid approaches begin to connect LLMs with collaborative signals (e.g., CoLLM \cite{zhang2025collm}; LLM-CF \cite{sun2024large}), yet a principled alignment of language priors with graph-derived embeddings, paired with lightweight, deployable fusion, remains underexplored. RecMind targets this gap via contrastive alignment between LLM-conditioned and graph-conditioned representations, followed by efficient fusion in a GNN backbone.

\section{Problem Formulation}
Let $\mathcal{U}$ be the set of users and $\mathcal{I}$ the set of items. We observe an implicit-feedback interaction graph $\mathcal{G}=(\mathcal{U}\cup\mathcal{I}, \mathcal{E})$, where $(u,i)\in\mathcal{E}$ indicates that user $u$ interacted with item $i$, e.g., view/click/purchase. For each node we have structured features $x_u$, $x_i$, and text metadata: user-side textual signals $T_u$, e.g., reviews, queries and item-side textual signals $T_i$, e.g., title, description, and attributes.

For a target user $u$, produce a top-$K$ ranked list over candidate items $\mathcal{I}\setminus \mathcal{I}_u^{+}$ that maximizes ranking utility (NDCG@K, Recall@K), with strong performance in (1) sparse regimes and (2) cold-start cases where $x_i$, $T_i$ are available but $(\cdot,i)\notin\mathcal{E}$.

\section{Methodology}
\begin{figure*}[htbp]
    \centering
    \includegraphics[width=\linewidth]{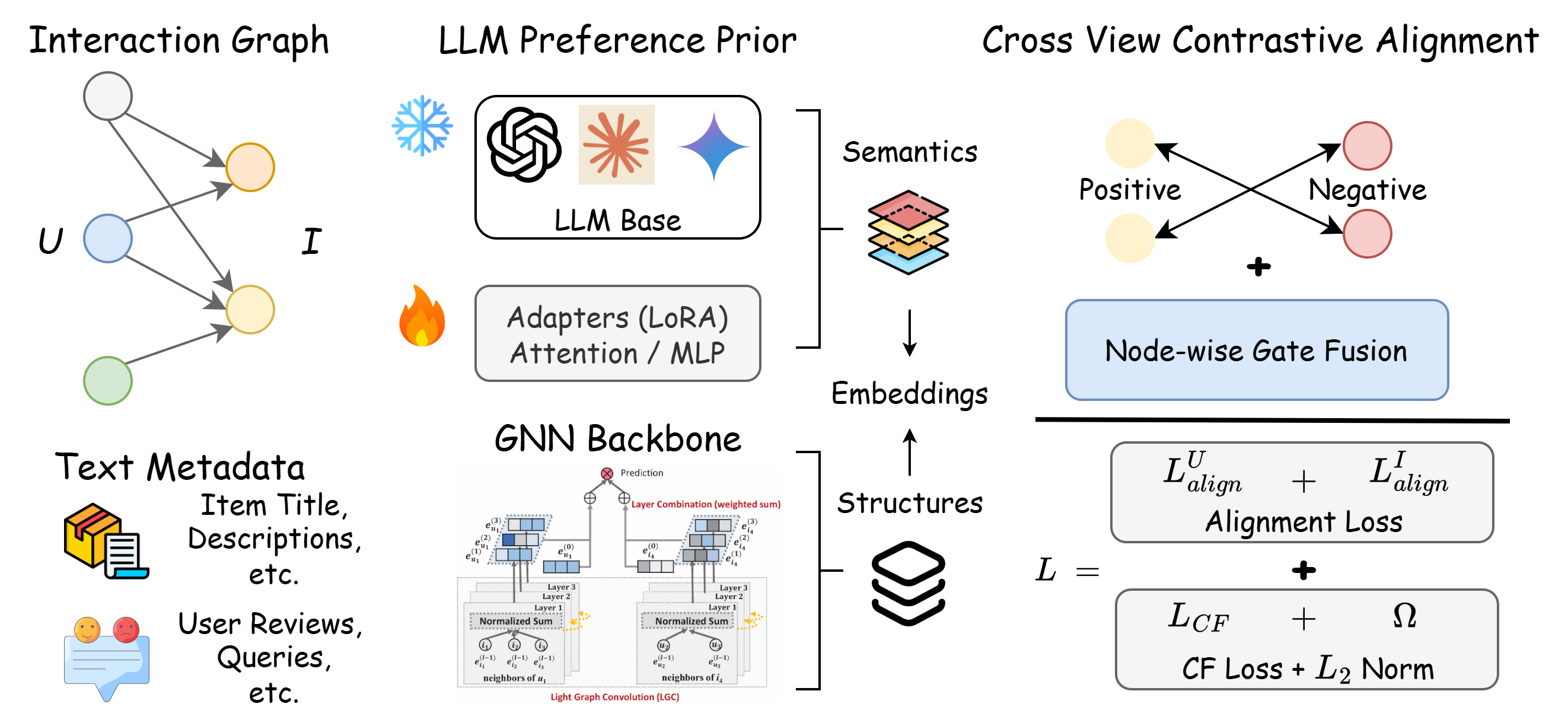}
    \caption{RecMind architecture: (1) LLM with adapters encodes text metadata into language embeddings; (2) GNN encodes interaction graph into graph embeddings; (3) contrastive loss aligns the two views; (4) fusion gate integrates them during message passing for final scoring.}
    \label{fig:architecture}
\end{figure*}

RecMind couples a graph-based collaborative encoder with a language-driven preference prior. Given an interaction graph $\mathcal{G}$ and text metadata $(T_u, T_i)$, we produce two embeddings per entity $v \in \mathcal{U}\cup\mathcal{I}$: a graph embedding $z^{G}_v$ from a GNN and a language embedding $z^{L}_v$ from a frozen LLM with lightweight adapters. We then (1) \textbf{align} these views via a contrastive loss so that textual semantics and collaborative structure agree, and (2) \textbf{fuse} them with a learned gate inside the message-passing pathway to yield final representations $h_v$ for scoring. Figure \ref{fig:architecture} outlines the architecture.

\subsection{GNN Backbone}
We adopt a LightGCN-style encoder for implicit-feedback recommendation due to its accuracy/efficiency trade-off \cite{he2020lightgcn}. Let ${E}^{(0)} \in \mathbb{R}^{|\mathcal{U}\cup\mathcal{I}| \times d}$ be trainable base embeddings. Message passing uses the normalized bipartite adjacency $\hat{{A}}$:
\begin{equation}
    {E}^{(l+1)}=\hat{{A}}\,{E}^{(l)},\quad l=0,\dots,L-1,
\end{equation}
and the graph embedding is the layer-averaged representation.

\begin{equation}
    z^{G}_v=\frac{1}{L+1}\sum_{l=0}^{L}{E}^{(l)}_v.
\end{equation}

This backbone captures high-order connectivity without nonlinearities or per-layer weight matrices, keeping computation $O(|\mathcal{E}|d)$ per layer and enabling easy deployment on large catalogs. We train the CF component with a standard pairwise ranking objective BPR over scores $s(u,i)=\langle h_u, h_i\rangle$. Dropout on edges or layer-wise embedding dropout can be used to regularize over-smoothing \cite{xing2024multi}.

\subsection{LLM Preference Module}
\textbf{Text construction.} For each user $u$, we concatenate up to $M_u$ short snippets (recent reviews/queries) with lightweight prompts; for each item $i$, we compose title, attributes, and a trimmed description. We normalize punctuation, remove boilerplate, and truncate to the LLM's token limit.

\textbf{Frozen LLM + adapters.} We treat the LLM $\mathcal{F}$ as a frozen encoder and attach parameter-efficient adapters LoRA only on attention/MLP blocks to condition on consumer metadata while keeping latency and memory modest \cite{hu2022lora}. Given text $T_v$, we extract the hidden state at a special pooling token and project to the GNN dimension:
\begin{equation}
    \tilde{z}^{L}_v=\mathrm{Pool}\left(\mathcal{F}(T_v;\text{adapters})\right)
\end{equation}

\begin{equation}
    z^{L}_v={W}_{\text{proj}}\tilde{z}^{L}_v \in \mathbb{R}^{d}
\end{equation}

Adapters and ${W}_{\text{proj}}$ are trainable; the LLM weights are frozen. This yields semantically rich, stable representations that can be cached offline for items and periodically refreshed for users. When text is missing, we fall back to structured features encoded through a small MLP into the same space.

\subsection{Cross-Modal Alignment and Fusion}
We align the graph and language views so that they agree on each entity while remaining discriminative across entities. For users (similarly for items), we define temperature-scaled InfoNCE:
\begin{equation}
    \mathcal{L}^{U}_{\text{align}} =-\frac{1}{|\mathcal{B}_U|}\sum_{u\in\mathcal{B}_U} \log \frac{\exp\!\left(\cos\!\left(z^{G}_u,z^{L}_u\right)/\tau\right)} {\sum_{u' \in \mathcal{B}_U}\exp\!\left(\cos\!\left(z^{G}_u,z^{L}_{u'}\right)/\tau\right)}
\end{equation}

We use a symmetric form and add the item term $\mathcal{L}^{I}_{\text{align}}$. Hard negatives are obtained by in-batch sampling. We maintain a small queue (momentum-updated keys) for more diverse negatives without extra forward passes \cite{xing2024enhancing}. This alignment produces language-aware graph embeddings and structure-aware language embeddings, which improves generalization under sparsity/cold-start.

\subsection{Fusion Inside Message Passing}
Rather than post-hoc concatenation, RecMind inserts a gating fusion that adapts the contribution of $z^{G}_v$ vs. $z^{L}_v$ per node and layer, letting language dominate when graph evidence is weak (e.g., cold items) and vice versa.

\textbf{Node-wise gate.} We compute a scalar gate $\gamma^{(l)}_v \in [0,1]$ using a small MLP over the concatenation of current representations and a normalized degree feature:
\begin{equation}
    \gamma^{(l)}_v=\sigma\!\left({w}^\top\!\left[{E}^{(l)}_v \,\|\, z^{L}_v \,\|\, \tilde{d}_v\right]+b\right),\quad \tilde{d}_v=\log(1+\mathrm{deg}(v))/c.
\end{equation}

The fused state used for propagation is
\begin{equation}
    \hat{{E}}^{(l)}_v = \gamma^{(l)}_v\,{E}^{(l)}_v + \left(1-\gamma^{(l)}_v\right)z^{L}_v
\end{equation}

We then pass $\hat{{E}}^{(l)}$ through the same LightGCN update:
\begin{equation}
    {E}^{(l+1)}=\hat{{A}}\,\hat{{E}}^{(l)}
\end{equation}

Finally, we average layers to obtain$z^{G}_v$ (now language-aware by construction) and define the final representation
\begin{equation}
    h_v=\alpha\, z^{G}_v + (1-\alpha)\, z^{L}_v
\end{equation}
where $\alpha$ is a learned global scalar. Scores are $s(u,i)=\langle h_u,h_i\rangle$.

\subsection{Training Objectives}
The full loss combines CF ranking and alignment:
\begin{equation}
    \mathcal{L} =\underbrace{\mathcal{L}_{\text{CF}}}_{\text{BPR on}\ s(u,i)} +\lambda\left(\mathcal{L}^{U}_{\text{align}}+\mathcal{L}^{I}_{\text{align}}\right) +\beta\,\Omega
\end{equation}
with $\Omega$ including $L_2$ regularization on embeddings/gates and adapter weight decay. We use mini-batches of users with one positive and $n$ sampled negatives; negatives are popularity-aware unless stated otherwise. A two-phase schedule improves stability: (1) warm-up for $T_w$ epochs optimizing only $\mathcal{L}_{\text{align}}$  to place both views in a shared space; (2) joint training of CF + alignment while updating base embeddings, gates, and adapters. Early stopping monitors validation NDCG@10.

\section{Experiments}
\subsection{Datasets}
We evaluate on three consumer-focused, text-rich benchmarks to stress both collaborative and semantic signals.
(1) Amazon-Electronics: product titles, attributes, and reviews paired with implicit interactions (purchases/clicks).
(2) Yelp: business metadata (name, categories, attributes) and user reviews with user-business interactions.

For each dataset, we apply a core-5 filter (users/items with $\geq$5 interactions), lowercase/normalize text, strip boilerplate, and keep field-aware token budgets (e.g., 32 tokens for titles, 96 for descriptions, 64 for attributes/reviews). We build a chronological leave-one-out split per user: last item for test, second-to-last for validation, the rest for training. To probe robustness, we define cold-start subsets by degree thresholds (e.g., items with degree $\leq$3; new items unseen in training when available).

\begin{table*}[htbp]
    \centering
    \caption{Overall performance on Yelp and Amazon-Electronics.\\RecMind consistently outperforms strong baselines across metrics and datasets.}
    \begin{tabular}{lcccccccc}
        \toprule
        \multicolumn{1}{c}{\multirow{2}[4]{*}{Baseline}} & \multicolumn{4}{c}{Yelp}                                     & \multicolumn{4}{c}{Amazon-Electronics}                                                                                                                         \\
        \cmidrule(lr){2-5} \cmidrule(lr){6-9}            & Recall@20                                                    & Recall@40                              & NDCG@20           & NDCG@40           & Recall@20         & Recall@40         & NDCG@20           & NDCG@40           \\
        \midrule
        \midrule
                                                         & \multicolumn{8}{c}{General Collaborative Filtering Methods}                                                                                                                                                                   \\
        \midrule
        BPR-MF \cite{rendle2012bpr}                      & \uline{0.1226}                                               & 0.1461                                 & 0.1003            & 0.1114            & 0.1235            & \uline{0.1811}    & 0.0717            & 0.1047            \\
        LightGCN \cite{he2020lightgcn}                   & 0.0909                                                       & 0.1605                                 & \uline{0.1128}    & 0.1132            & 0.1174            & 0.1637            & 0.0841            & 0.0960            \\
        \midrule
                                                         & \multicolumn{8}{c}{Recommenders with Sequential Information}                                                                                                                                                                  \\
        \midrule
        SASRec \cite{kang2018self}                       & 0.1076                                                       & 0.1455                                 & 0.1018            & 0.1031            & \uline{0.1355}    & 0.1802            & 0.0822            & 0.0969            \\
        BERT4Rec \cite{sun2019bert4rec}                  & 0.1157                                                       & 0.1569                                 & 0.0718            & 0.0919            & 0.1207            & 0.1610            & 0.0634            & 0.1152            \\
        \midrule
                                                         & \multicolumn{8}{c}{LLM-augmented Methods}                                                                                                                                                                                     \\
        \midrule
        LLMRec \cite{wei2024llmrec}                      & 0.1174                                                       & 0.1415                                 & 0.0893            & \uline{0.1176}    & 0.1169            & 0.1751            & \uline{0.0876}    & 0.1016            \\
        SAID \cite{hu2024enhancing}                      & 0.1040                                                       & \uline{0.1682}                         & 0.0709            & 0.0986            & 0.1213            & 0.1584            & 0.0755            & \uline{0.1154}    \\
        \midrule
        RecMind                                          & \textbf{0.1259}                                              & \textbf{0.1741}                        & \textbf{0.1166}   & \textbf{0.1223}   & \textbf{0.1385}   & \textbf{0.1893}   & \textbf{0.0880}   & \textbf{0.1180}   \\
        Improve                                          & 2.69\% $\uparrow$                                            & 3.53\% $\uparrow$                      & 3.36\% $\uparrow$ & 4.01\% $\uparrow$ & 2.24\% $\uparrow$ & 4.53\% $\uparrow$ & 0.47\% $\uparrow$ & 2.28\% $\uparrow$ \\
        \bottomrule
    \end{tabular}%
    \label{tab:overall performance}
\end{table*}%

\subsection{Baselines}
We compare RecMind to strong families of recommenders: (1) General CF: \textbf{BPR-MF} \cite{rendle2012bpr} and \textbf{LightGCN} \cite{he2020lightgcn} (best-performing GNN CF backbone). (2) Sequential: \textbf{SASRec} \cite{kang2018self} and \textbf{BERT4Rec} \cite{sun2019bert4rec} for order-aware next-item prediction. (3) LLM-augmented: \textbf{LLMRec} \cite{wei2024llmrec} and \textbf{SAID} \cite{hu2024enhancing}.

All baselines receive the same candidate pool, splits, and negatives. Hyperparameters are tuned on validation via grid search (embedding sizes, layers, dropout, learning rates); we adopt each method's recommended settings when available to avoid under-tuning.

\subsection{Evaluation Metrics}
We report Recall@K and NDCG@K (K = 20, 40) using all-item ranking with 100 sampled negatives per user unless noted; the positive is withheld chronologically. Metrics are averaged over users; higher is better. We additionally stratify results by item/user degree to quantify cold-start behavior and report relative improvement (\%) over the strongest baseline.

\subsection{Results}
Table \ref{tab:overall performance} reports top-K ranking on Yelp and Amazon-Electronics. RecMind achieves the best score on all eight metrics. On Yelp, it reaches Recall@20/40 = 0.1259/0.1741 and NDCG@20/40 = 0.1166/0.1223, yielding relative gains of +2.69\%, +3.53\%, +3.36\%, and +4.01\% over the strongest baseline, respectively. On Amazon-Electronics, RecMind delivers Recall@20/40 = 0.1385/0.1893 and NDCG@20/40 = 0.0880/0.1180, improving by +2.24\%, +4.53\%, +0.47\%, and +2.28\%. Improvements are consistent across recall and discount-sensitive metrics, indicating better ranking both at the top and deeper in the list.

Gains over BPR-MF/LightGCN and SASRec/BERT4Rec confirm that collaborative structure or sequence signals alone under-exploit rich consumer text. The largest relative boosts (e.g., +4.01\% NDCG@40 on Yelp and +4.53\% Recall@40 on Amazon) suggest RecMind retrieves more relevant tail items while re-ranking them more effectively.

RecMind also surpasses prompt-based and late-fusion LLM methods, highlighting the benefit of (1) contrastive alignment between language and graph derived views and (2) intra-layer gating that adapts to data sparsity. The smallest margin (+0.47\% NDCG@20 on Amazon) is expected where metadata is concise and interaction structure dominates; even there, alignment/fusion prevents LLM drift without hurting precision.

Across both domains, the pattern matches our design goals: semantics help cold/long-tail retrieval, while graph constraints maintain stability and precision.

\begin{table*}[htbp]
    \centering
    \caption{Ablation study on Yelp and Amazon-Electronics.\\Removing alignment or LLM degrades performance, confirming the importance of both components.}
    \begin{tabular}{lcccccccc}
        \toprule
        \multicolumn{1}{c}{\multirow{2}[4]{*}{Baseline}} & \multicolumn{4}{c}{Yelp} & \multicolumn{4}{c}{Amazon-Electronics}                                                                                                             \\
        \cmidrule(lr){2-5} \cmidrule(lr){6-9}            & Recall@20                & Recall@40                              & NDCG@20         & NDCG@40         & Recall@20       & Recall@40       & NDCG@20         & NDCG@40         \\
        \midrule
        LLM-only                                         & 0.1141                   & 0.1540                                 & 0.1095          & 0.1164          & 0.1198          & 0.1876          & 0.0798          & 0.1115          \\
        w/o Alignment-Users                              & 0.1079                   & 0.1468                                 & 0.1053          & 0.1187          & 0.1176          & 0.1761          & 0.0828          & 0.1006          \\
        w/o Alignment-Items                              & 0.0974                   & 0.1570                                 & 0.1090          & 0.1194          & 0.1160          & 0.1789          & 0.0801          & 0.0991          \\
        RecMind                                          & \textbf{0.1259}          & \textbf{0.1741}                        & \textbf{0.1166} & \textbf{0.1223} & \textbf{0.1385} & \textbf{0.1893} & \textbf{0.0880} & \textbf{0.1180} \\
        \bottomrule
    \end{tabular}%
    \label{tab:ablation study}%
\end{table*}%

\subsection{Ablation Study}
We ablate RecMind to isolate the effect of its main components: (1) LLM-only, a prompt-based ranker that scores candidates with a frozen LLM (no graph signals), and (2) w/o Alignment-Users and w/o Alignment-Items, which remove the corresponding cross-view contrastive terms. Results are shown in Table \ref{tab:ablation study}.

Using the LLM alone consistently underperforms the full model. On Yelp, Recall@20 drop from 0.1259 to 0.1141 (-9.4\%), and NDCG@20 fall to 0.1095 (-6.1\%). Similar results can be observed when K is set to 40. On Amazon-Electronics, Recall@20 decreases from 0.1385 to 0.1198 (-13.5\%) and NDCG@20 from 0.0880 to 0.0798 (-9.3\%); Recall@40 is close (0.1893 $\rightarrow$ 0.1876), but top-rank precision still degrades (NDCG@40: 0.1180 $\rightarrow$ 0.1115). This confirms that semantic cues alone are insufficient without collaborative structure.

Removing either alignment term hurts across datasets, validating our design. On Yelp, w/o Alignment-Items is most damaging to Recall@20 (0.0974, -22.7\%), indicating item-side language signals require alignment to translate into better retrieval. On Amazon, both terms are important for deep ranking: NDCG@40 drops to 0.1006 (-14.7\%) without user alignment and to 0.0991 (-16.0\%) without item alignment. Overall, the full RecMind variant benefits from both alignment and graph-language fusion, especially for top-K precision and cold/long-tail retrieval.

\section{Conclusion}
We introduced RecMind, a practical architecture that unifies GNN-based collaborative reasoning with LLM-driven semantic priors through contrastive alignment and intra-layer gated fusion. Experiments on two consumer datasets show consistent gains in HR@K/NDCG@K and particularly strong improvements on cold/long-tail items, while ablations validate the contribution of each component. Because the LLM is frozen and adapters are small, RecMind remains deployable with precomputed item representations and modest online overhead, properties aligned with real-world consumer systems. Limitations include dependence on text quality, token budgets for long descriptions, and possible domain shift across product categories. Future work will explore instruction-tuned language priors for recommendation, bias/fairness auditing and explanation generation, and joint training with re-ranking to further improve reliability and user trust.

\bibliographystyle{IEEEtran}
\bibliography{main}
\end{document}